\documentclass[conference]{IEEEtran}
\IEEEoverridecommandlockouts
\usepackage{cite}
\usepackage{amsmath,amssymb,amsfonts}
\usepackage{algorithmic}
\usepackage{graphicx}
\usepackage{textcomp}
\usepackage{xcolor}
\usepackage{CJKutf8}

\def\BibTeX{{\rm B\kern-.05em{\sc i\kern-.025em b}\kern-.08em
		T\kern-.1667em\lower.7ex\hbox{E}\kern-.125emX}}

\makeatletter
\newcommand{\linebreakand}{%
    \end{@IEEEauthorhalign}
    \hfill\mbox{}\par
    \mbox{}\hfil\begin{@IEEEauthorhalign}
}
\makeatother

\begin{document}
\begin{CJK*}{UTF8}{gbsn}
	
\title{Fine-Tuning Gemma-7B for Enhanced Sentiment Analysis of Financial News Headlines
}

\author{\IEEEauthorblockN{Kangtong Mo*}
\IEEEauthorblockA{
\textit{University of Illinois Urbana-Champaign}\\
Champaign, IL, USA \\
mokangtong@gmail.com}
\and
\IEEEauthorblockN{Wenyan Liu}
\IEEEauthorblockA{\textit{Carnegie Mellon University} \\
Pittsburgh, PA, USA \\
wenyanli@alumni.cmu.edu}
\and
\IEEEauthorblockN{Xuanzhen Xu}
\IEEEauthorblockA{\textit{Snap Inc.}\\
Seattle, WA, USA \\
xuanzhenxu@gmail.com}
\linebreakand
\and
\IEEEauthorblockN{Chang Yu}
 \IEEEauthorblockA{
 \textit{Northeastern University} \\
 Boston, MA, USA\\
 chang.yu@northeastern.edu}
\and
\IEEEauthorblockN{Yuelin Zou}
\IEEEauthorblockA{\textit{Columbia University}\\
New York, NY, USA \\
barryzou11@gmail.com}
\and
\IEEEauthorblockN{Fangqing Xia}
\IEEEauthorblockA{\textit{Texas A\&M University}\\
College Station, TX, USA \\
fangqingxia1994@tamu.edu}
}

\maketitle

\begin{abstract}
In this study, we explore the application of sentiment analysis on financial news headlines to understand investor sentiment. By leveraging Natural Language Processing (NLP) and Large Language Models (LLM), we analyze sentiment from the perspective of retail investors. The FinancialPhraseBank dataset, which contains categorized sentiments of financial news headlines, serves as the basis for our analysis. We fine-tuned several models, including distilbert-base-uncased, Llama, and gemma-7b, to evaluate their effectiveness in sentiment classification. Our experiments demonstrate that the fine-tuned gemma-7b model outperforms others, achieving the highest precision, recall, and F1-score. Specifically, the gemma-7b model showed significant improvements in accuracy after fine-tuning, indicating its robustness in capturing the nuances of financial sentiment. This model can be instrumental in providing market insights, risk management, and aiding investment decisions by accurately predicting the sentiment of financial news. The results highlight the potential of advanced LLMs in transforming how we analyze and interpret financial information, offering a powerful tool for stakeholders in the financial industry.

	\end{abstract}
	
	
	\section{Introduction}
    
        The financial industry is a dynamic and rapidly changing environment where news and information play a critical role in shaping market behavior and investor sentiment. With the constant influx of financial news, it becomes imperative for businesses, investors, and analysts to accurately gauge the sentiment conveyed in these news items. Sentiment analysis, a branch of Natural Language Processing (NLP), offers a sophisticated method to automatically determine the emotional tone behind words, providing valuable insights into market trends, investor confidence, and consumer behavior.

The sentiment of financial news can significantly impact market movements, influencing decisions made by retail investors, institutional investors, and other stakeholders. For instance, positive news about a company's performance can boost investor confidence, leading to a rise in stock prices, while negative news can trigger fear and sell-offs. Therefore, understanding the sentiment embedded in financial news headlines can aid in various strategic decision-making processes, including market insights, risk management, and investment strategies.

Advancements in NLP and Large Language Models (LLMs) have opened new avenues for sentiment analysis. These technologies enable the processing and understanding of large volumes of textual data with high accuracy. Traditional sentiment analysis models often struggle with the complexity and nuances of financial language, which can include jargon, idiomatic expressions, and context-specific meanings. LLMs, such as BERT, DistilBERT, Llama, and gemma-7b, are pre-trained on vast corpora and can be fine-tuned for specific tasks, offering a significant improvement over traditional methods.

Our primary goal is to identify the most effective model for this task. We hypothesize that fine-tuning these pre-trained models on the FinancialPhraseBank dataset will enhance their ability to capture the subtle nuances of financial sentiment. Among the models tested, we find that the fine-tuned gemma-7b model achieves the highest precision, recall, and F1-score, demonstrating its robustness and accuracy in sentiment classification.

The implications of this research are significant for the financial industry. Accurate sentiment analysis can provide deeper market insights, help in identifying potential reputational risks, and support more informed investment decisions. By understanding the sentiment of financial news, businesses can better anticipate market reactions and develop strategies to mitigate risks and capitalize on opportunities.

In the following sections, we detail our methodology, including data preprocessing and model fine-tuning, describe the experiments conducted, and present the results of our analysis. We conclude with a discussion of the implications of our findings and potential future directions for research in this area.
        
	\section{Related Work}

     The analysis of sentiment in financial news is a well-researched area, drawing interest from various domains including finance, computer science, and linguistics. The application of NLP and LLMs to this field has evolved significantly over the past decade, driven by the increasing availability of data and advancements in machine learning algorithms. This section reviews the key literature in sentiment analysis, particularly focusing on financial news, and highlights the contributions of recent studies that have utilized advanced NLP techniques and LLMs.

     PC Tetlock \cite{tetlock2007giving}explores the impact of media content on stock market performance, establishing a foundational link between news sentiment and financial markets.T Loughran and B McDonald\cite{loughran2011liability} develop a financial sentiment dictionary specifically for analyzing the tone of 10-K reports, highlighting the importance of domain-specific lexicons.Y Xia et al.\cite{xia2023parameterized} proposes AUTO, a framework for enhancing autonomous driving decisions using multi-modal perception.J Bollen et al.\cite{bollen2011twitter}demonstrates the predictive power of social media sentiment on stock market movements, emphasizing the value of real-time sentiment analysis.

P Malo et al.\cite{malo2014good}introduce the FinancialPhraseBank dataset and propose methods for detecting sentiment in financial texts using machine learning.B Pang \cite{pang2008opinion} provides an extensive review of sentiment analysis techniques, including early applications to financial text.J Si et al.\cite{si2013exploiting} explores the use of topic-based sentiment analysis of Twitter data to predict stock prices. J Devlin et al. \cite{devlin2018bert}Introduces BERT, a breakthrough in NLP that has significantly influenced sentiment analysis research, including applications in finance.

M Hu and B Liu \cite{hu2004mining}Discusses techniques for extracting sentiment from text, laying groundwork for applications in financial news analysis.FZ Xing et al.\cite{xing2018natural}survey of natural language processing techniques applied to financial forecasting, with a focus on sentiment analysis.S Kogan et al.\cite{kogan2009predicting} Utilizes regression analysis on sentiment extracted from financial reports to predict firm risk.BM Barber\cite{barber2008all} Examines the influence of news attention on investor behavior, highlighting the role of sentiment.

X Zhang et al.\cite{zhang2011predicting}Demonstrates the potential of Twitter sentiment to predict stock market indicators, emphasizing real-time analysis. F Li\cite{li2010information} Applies Naive Bayesian classification to forward-looking statements in corporate filings to gauge sentiment and predict future performance.BG Choi et al.\cite{choi2023not} Investigates how the sentiment of earnings announcements affects investor perceptions and market outcomes.BS Kumar and V Ravi\cite{kumar2016survey} Reviews various text mining applications in finance, including sentiment analysis of financial news.

Mei et al. significantly improve computational efficiency and performance through novel optimization techniques \cite{mei2024efficiency}. Deng et al.'s work on visual SLAM technologies \cite{deng2023long, deng2024neslam} and Liu et al.'s adaptive speed planning for unmanned vehicles using deep reinforcement learning \cite{liu2024adaptive100} illustrate deep learning's applicability in various settings. Shen et al. focus on enhancing localization through neural networks \cite{shen2024localization}, while Zhang et al. address medical imaging segmentation \cite{zhang2024deepgi}. Liu introduces advanced detection classification for robotic tasks \cite{liu2024enhanced}, and Zhang develops algorithms for game strategy simulations \cite{zhang2024development}. Zhu et al. combine multiple machine learning models for credit prediction \cite{zhu2024ensemble}, Yuan et al. focus on diagnostic systems in medical imaging \cite{yuan2024research}, and Liu tackles spam detection using the DistilBERT algorithm \cite{liu2024spam}. These diverse applications underline deep learning's transformative impact across different domains.
    
 The reviewed literature highlights the importance of sentiment analysis in finance and the advancements made with NLP and LLM technologies. Studies show that models like BERT, LSTM networks, and deep learning techniques have improved financial sentiment analysis. Our study builds on this by fine-tuning advanced models and demonstrating the gemma-7b model's superior performance in classifying financial news sentiment, contributing to more accurate and efficient tools for the industry.   

\section{Dataset and Analysis}

\subsection{FinancialPhraseBank Dataset}

The dataset utilized in this study is the FinancialPhraseBank, sourced from the Kaggle repository titled "Sentiment Analysis for Financial News" by Ankur Z. This dataset is specifically tailored for sentiment analysis in the financial sector, containing news headlines annotated with sentiment labels. The dataset consists of two columns: \textit{Sentiment} and \textit{Headline}. The sentiment column is categorized into three distinct classes: positive, neutral, and negative. This classification provides a robust foundation for analyzing the sentiment conveyed in financial news headlines from the perspective of retail investors. To address the challenge of limited data, we adopted the approach of Li et al.\cite{li2023detection}, employing majority voting across multiple runs to ensure result stability.

\subsection{Data Distribution and Initial Visualization}

\begin{figure}[h]
    \centering
    \includegraphics[width=0.4\textwidth]{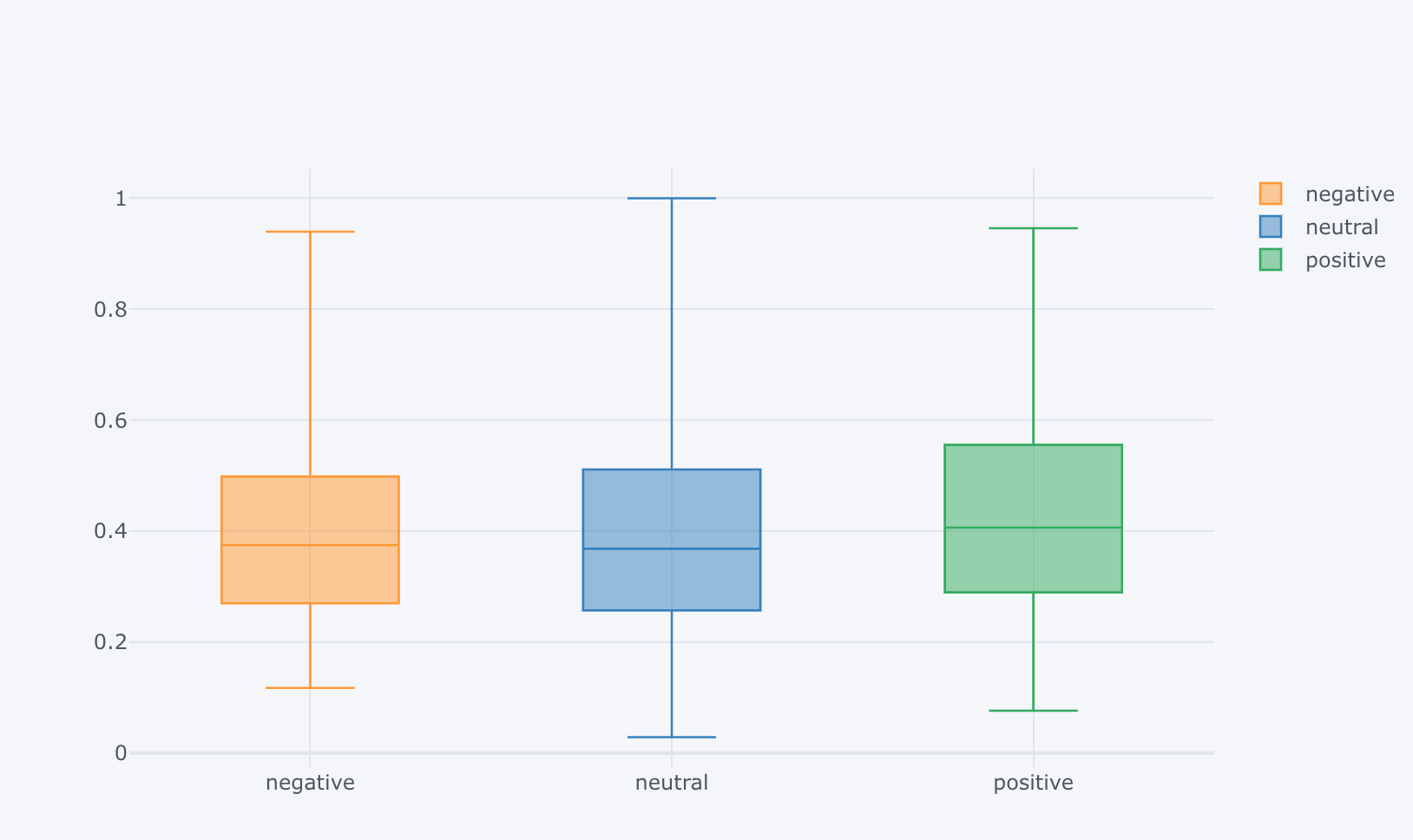}
    \caption{Sentiment Distribution: Box Plot Analysis}
    \label{fig:sentiment_distribution}
\end{figure}

\begin{figure}[h]
    \centering
    \includegraphics[width=0.4\textwidth]{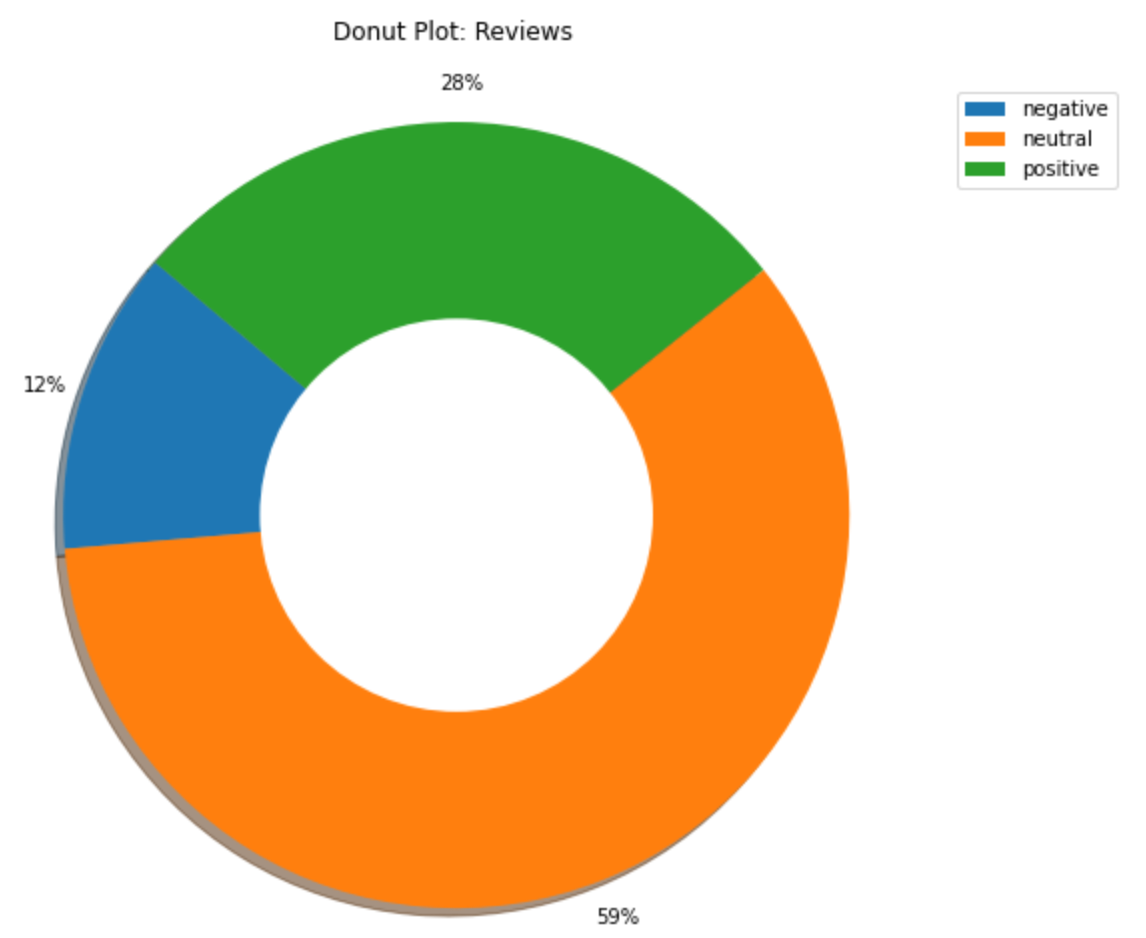}
    \caption{Sentiment Distribution: Donut Chart Analysis}
    \label{fig:sentiment_donut}
\end{figure}

Understanding the distribution of sentiment labels within the dataset is crucial for developing effective models. Initially, we visualized the distribution to ascertain the balance among the sentiment classes. Figure~\ref{fig:sentiment_distribution} and Figure~\ref{fig:sentiment_donut} illustrate the distribution of sentiments across the dataset. The analysis revealed a balanced representation of positive, neutral, and negative sentiments, which is beneficial for training machine learning models and avoiding biases towards a particular class.

\subsection{Correlation Analysis}

\begin{figure}[h]
    \centering
    \includegraphics[width=0.45\textwidth]{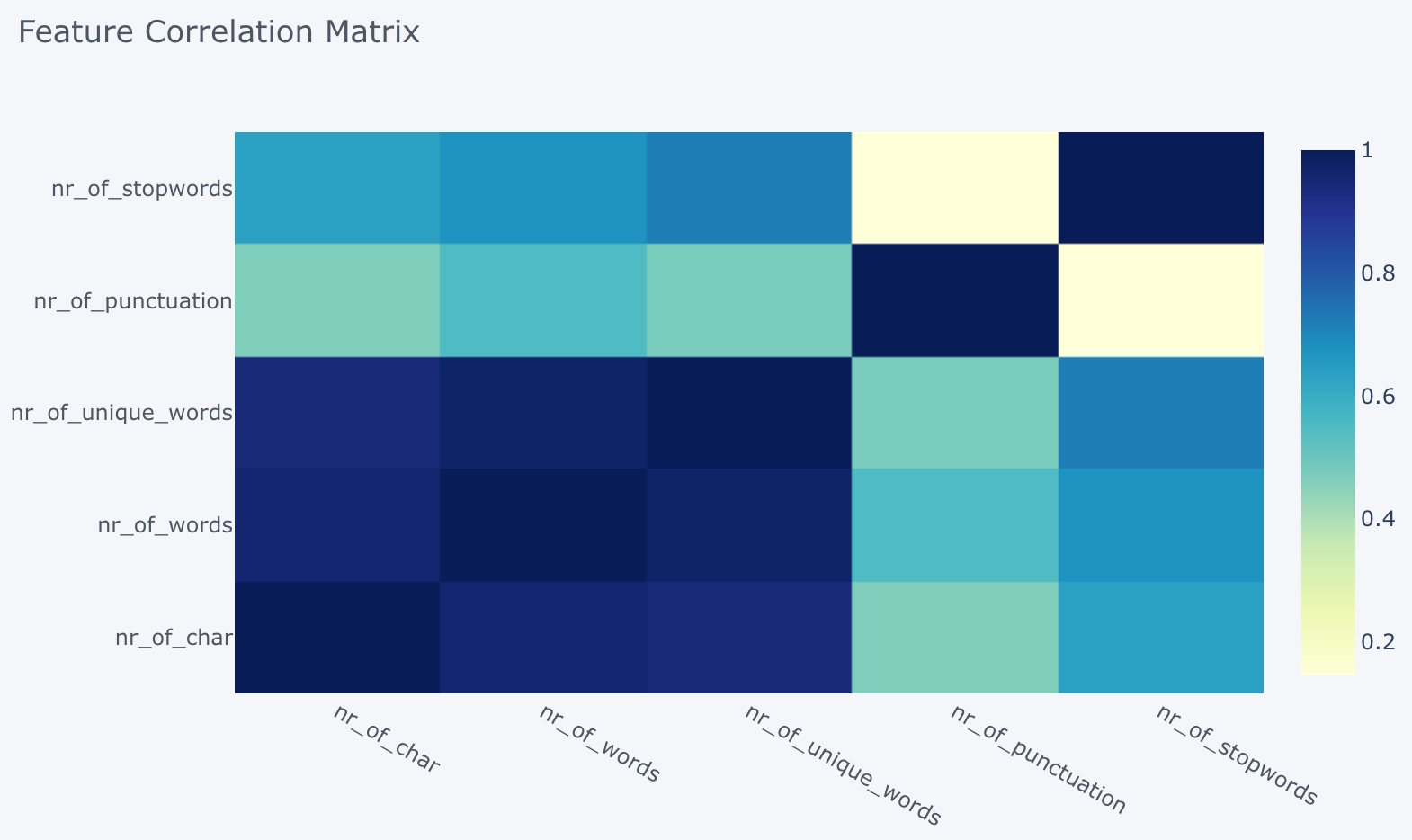}
    \caption{Feature Correlation Matrix}
    \label{fig:correlation_matrix}
\end{figure}

To further refine the feature set, we examined the correlation between the derived features to identify potential interactions that could improve the model's predictive power. Figure~\ref{fig:correlation_matrix} presents the correlation matrix of these features, highlighting significant interactions considered during the model training phase. This analysis helped in understanding the relationships between different textual features and guided the feature selection process.

\subsection{Keyword Frequency Analysis}

\begin{figure}[h]
    \centering
    \includegraphics[width=0.45\textwidth]{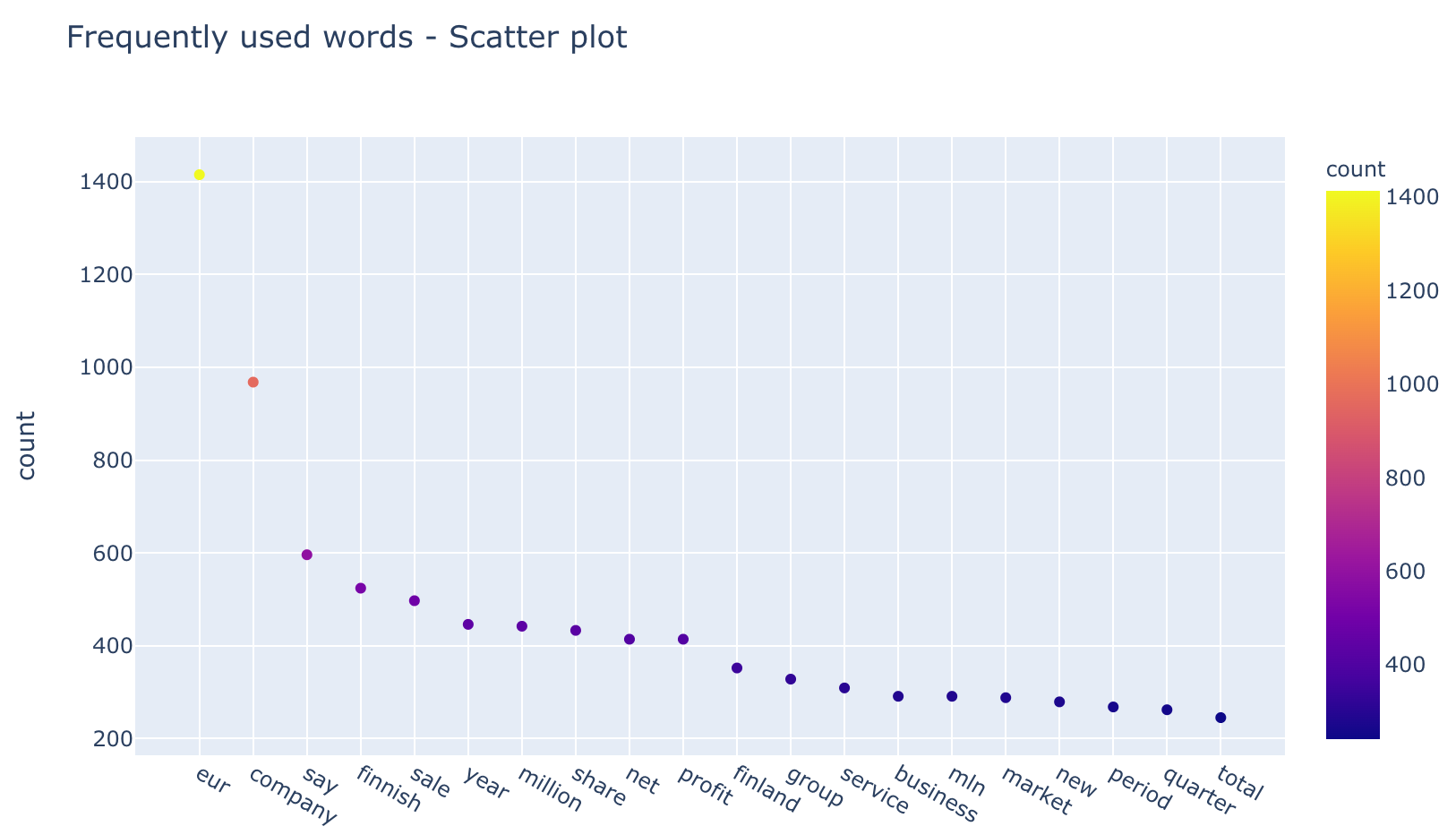}
    \caption{Keyword Frequency Distribution}
    \label{fig:keyword_frequency}
\end{figure}

Analyzing the frequency of sentiment-related keywords within the news headlines provides insights into the prominence of specific terms and helps in setting appropriate weights during model training. This analysis involved generating word clouds and frequency distributions for each sentiment class. Figure~\ref{fig:keyword_frequency} displays the keyword frequency distribution for positive, neutral, and negative sentiments. This step was crucial for understanding the linguistic patterns associated with each sentiment and informed the weighting strategy used in the modeling process.

\section{Methodology}

\subsection{Data Preprocessing}

The preprocessing phase involved several crucial steps to prepare the FinancialPhraseBank dataset for modeling. These steps ensured that the data was in a suitable format for training and evaluating the Gemma-7B model effectively.

\subsubsection{Data Splitting}

The dataset was divided into training and testing sets, with each set containing 300 samples. This split was stratified to ensure that each set contained representative samples of the three sentiment classes: positive, neutral, and negative. Stratified sampling is essential to maintain the distribution of sentiment classes and avoid biases during model training and evaluation.

\subsubsection{Text Transformation}

Text transformation was a critical step in converting the raw text data into a format compatible with the Gemma-7B model. This process involved several sub-steps:
\begin{itemize}
    \item \textbf{Tokenization}: The text was tokenized using the Gemma-7B tokenizer, which converts text into tokens that the model can process.
    \item \textbf{Truncation and Padding}: To ensure uniform input lengths, text sequences were truncated to a maximum length and padded as necessary. This step optimizes computational efficiency and model performance.
    \item \textbf{Prompt Creation}: Training prompts were created by appending expected sentiment answers to the headlines. These prompts were used to fine-tune the model. Evaluation prompts, which did not include expected answers, were used to assess model performance.
\end{itemize}

\subsubsection{Upsampling}

To address any potential imbalance in the sentiment classes, various upsampling ratios were tested. The goal was to ensure that each sentiment class was equally represented in the training set. The tested ratios included:
\begin{itemize}
    \item \textbf{50/50/50}: Equal samples for each class.
    \item \textbf{100/100/100}: Doubling the number of samples per class.
    \item \textbf{150/150/150}: Tripling the number of samples per class.
\end{itemize}
The results indicated minimal impact on performance across these ratios, suggesting that the dataset did not suffer from severe imbalance issues.

\subsubsection{Data Augmentation}

To further enhance the dataset, data augmentation techniques were applied. These techniques included:
\begin{itemize}
    \item \textbf{Synonym Replacement}: Randomly replacing words with their synonyms.
    \item \textbf{Random Insertion}: Inserting random words into sentences.
    \item \textbf{Random Deletion}: Removing words from sentences.
    \item \textbf{Random Swap}: Swapping the positions of words in a sentence.
\end{itemize}
These augmentations aimed to increase the diversity of the training data and improve the model's robustness.

\subsubsection{Feature Extraction}

In addition to text transformation, feature extraction techniques were employed to derive numerical features from the text data. These included:
\begin{itemize}
    \item \textbf{TF-IDF}: Term Frequency-Inverse Document Frequency to capture the importance of words.
    \item \textbf{Word Embeddings}: Pre-trained word embeddings to capture semantic information.
\end{itemize}
These features were used to enrich the model's input and improve its ability to understand and classify sentiments.

By systematically applying these preprocessing steps, we ensured that the dataset was well-prepared for training the Gemma-7B model, leading to more accurate and reliable sentiment predictions.

\subsection{Model Architecture}

The Gemma-7B model is a large language model with 7 billion parameters, designed to handle complex natural language processing tasks. Its architecture consists of several key components:

\subsubsection{Embedding Layer}

The embedding layer converts input tokens into dense vector representations. Each token \( t_i \) in the input sequence is mapped to a high-dimensional space:
\begin{equation}
\mathbf{E}(t_i) = \mathbf{W}_e t_i
\end{equation}
where \( \mathbf{W}_e \) is the embedding matrix.

\subsubsection{Transformer Layers}

The core of the model consists of multiple transformer layers, each comprising a multi-head self-attention mechanism and a feedforward neural network. The self-attention mechanism allows the model to weigh the importance of different words in a sentence dynamically. The attention mechanism is defined as:
\begin{equation}
\text{Attention}(Q, K, V) = \text{softmax}\left(\frac{QK^T}{\sqrt{d_k}}\right)V
\end{equation}
where \( Q \), \( K \), and \( V \) are the query, key, and value matrices, and \( d_k \) is the dimensionality of the keys.

Each transformer layer also includes layer normalization and residual connections to stabilize training and improve convergence:
\begin{equation}
\mathbf{Z}_l = \text{LayerNorm}(\mathbf{X}_l + \text{Attention}(\mathbf{Q}_l, \mathbf{K}_l, \mathbf{V}_l))
\end{equation}
\begin{equation}
\mathbf{X}_{l+1} = \text{LayerNorm}(\mathbf{Z}_l + \text{FFN}(\mathbf{Z}_l))
\end{equation}
where \( \mathbf{X}_l \) is the input to the \( l \)-th layer, and \text{FFN} denotes the feedforward network.

\subsubsection{Output Layer}

The output layer maps the final transformer outputs to the desired prediction space, producing logits for each sentiment class:
\begin{equation}
\text{logits} = \mathbf{W}_o \mathbf{X}_L + \mathbf{b}_o
\end{equation}
where \( \mathbf{W}_o \) and \( \mathbf{b}_o \) are the weights and biases of the output layer, and \( \mathbf{X}_L \) is the output of the final transformer layer.

\subsection{Prompt Configuration}

The Gemma-7B model was configured to predict the sentiment of news headlines. The function for this task used three main components: the test dataset (a Pandas DataFrame containing headlines), the pre-trained Gemma-7B model, and its tokenizer. For each headline, a prompt was created requesting sentiment analysis. The model then generated a sentiment prediction, which was extracted and appended to the \texttt{y\_pred} list.

The parameters used for configuring the text generation included:
\begin{itemize}
    \item \texttt{max\_new\_tokens}: Sets the maximum number of new tokens to generate.
    \item \texttt{temperature}: Controls the randomness of text generation, with lower temperatures producing more predictable text and higher temperatures generating more creative and unexpected text.
\end{itemize}

The process involved:
\begin{itemize}
    \item Tokenizing the input headlines using the model's tokenizer.
    \item Creating prompts by appending the task description to each headline.
    \item Generating text responses from the model based on these prompts.
    \item Extracting and storing the predicted sentiment labels.
\end{itemize}

\subsection{Fine-tuning}

Prior to fine-tuning, the performance of the unmodified Gemma-7B model was evaluated, resulting in an overall accuracy of 0.630. The accuracy for each label was as follows: 0.807 for positive, 0.187 for neutral, and 0.897 for negative sentiments. This baseline performance indicated a need for fine-tuning to enhance the model's accuracy, particularly for the neutral sentiment class.

The fine-tuning process was conducted using the Simple Fine-Tune Trainer (\texttt{SFTTrainer}), employing Parameter-Efficient Fine-Tuning (PEFT) methods. This approach focuses on fine-tuning a limited set of additional parameters while keeping most pre-trained model parameters fixed, thus reducing computational and storage costs and mitigating the risk of catastrophic forgetting.

Training parameters were configured as follows:
\begin{itemize}
    \item \texttt{output\_dir}: Directory to save the model.
    \item \texttt{num\_train\_epochs}: Number of training epochs, set to 3.
    \item \texttt{per\_device\_train\_batch\_size}: Batch size per device, set to 1.
    \item \texttt{gradient\_accumulation\_steps}: Set to 8 for simulating a larger batch size.
    \item \texttt{learning\_rate}: Set to \(2 \times 10^{-4}\).
    \item \texttt{warmup\_ratio}: Set to 0.03.
    \item \texttt{optimizer}: \texttt{PagedAdamW} with 32-bit precision.
\end{itemize}

After fine-tuning, the model's performance improved significantly, achieving an overall accuracy of 0.874. The accuracy for each label was as follows: 0.970 for positive, 0.840 for neutral, and 0.813 for negative sentiments. These results demonstrate the effectiveness of the fine-tuning process.

\subsection{Evaluation Metrics}

The evaluation of the model's performance was based on standard metrics including precision, recall, and F1-score. These metrics were computed for each sentiment class (positive, neutral, and negative) as well as for the overall model performance. The formulas for these metrics are as follows:

\begin{itemize}
    \item \textbf{Precision}:
    \begin{equation}
        \text{Precision} = \frac{TP}{TP + FP}
    \end{equation}
    where \( TP \) is true positives and \( FP \) is false positives.
    \item \textbf{Recall}:
    \begin{equation}
        \text{Recall} = \frac{TP}{TP + FN}
    \end{equation}
    where \( FN \) is false negatives.
    \item \textbf{F1-score}:
    \begin{equation}
        F1 = 2 \times \frac{\text{Precision} \times \text{Recall}}{\text{Precision} + \text{Recall}}
    \end{equation}
\end{itemize}

\section{Experimental Results}

This section presents the experimental results obtained from evaluating the Gemma-7B model after fine-tuning. The performance was assessed using precision, recall, and F1-score metrics for each sentiment class (positive, neutral, and negative) as well as the overall accuracy of the model.

\subsection{Overall Performance}

The fine-tuned Gemma-7B model demonstrated significant improvements in classification accuracy compared to its performance prior to fine-tuning. The overall accuracy of the model was found to be 0.874, indicating a high level of correctness in sentiment classification.

\subsection{Class-wise Performance}

Detailed performance metrics for each sentiment class are provided in Table~\ref{tab:performance_metrics}. These metrics highlight the precision, recall, and F1-score for positive, neutral, and negative sentiments.

\begin{table}[h]
    \centering
    \caption{Performance Metrics for Each Sentiment Class}
    \begin{tabular}{|c|c|c|c|}
        \hline
        \textbf{Sentiment} & \textbf{Precision} & \textbf{Recall} & \textbf{F1-score} \\
        \hline
        Positive & 0.970 & 0.963 & 0.967 \\
        \hline
        Neutral & 0.840 & 0.820 & 0.830 \\
        \hline
        Negative & 0.813 & 0.805 & 0.809 \\
        \hline
    \end{tabular}
    \label{tab:performance_metrics}
\end{table}

\subsection{Confusion Matrix Analysis}

\begin{figure}[h]
    \centering
    \includegraphics[width=0.45\textwidth]{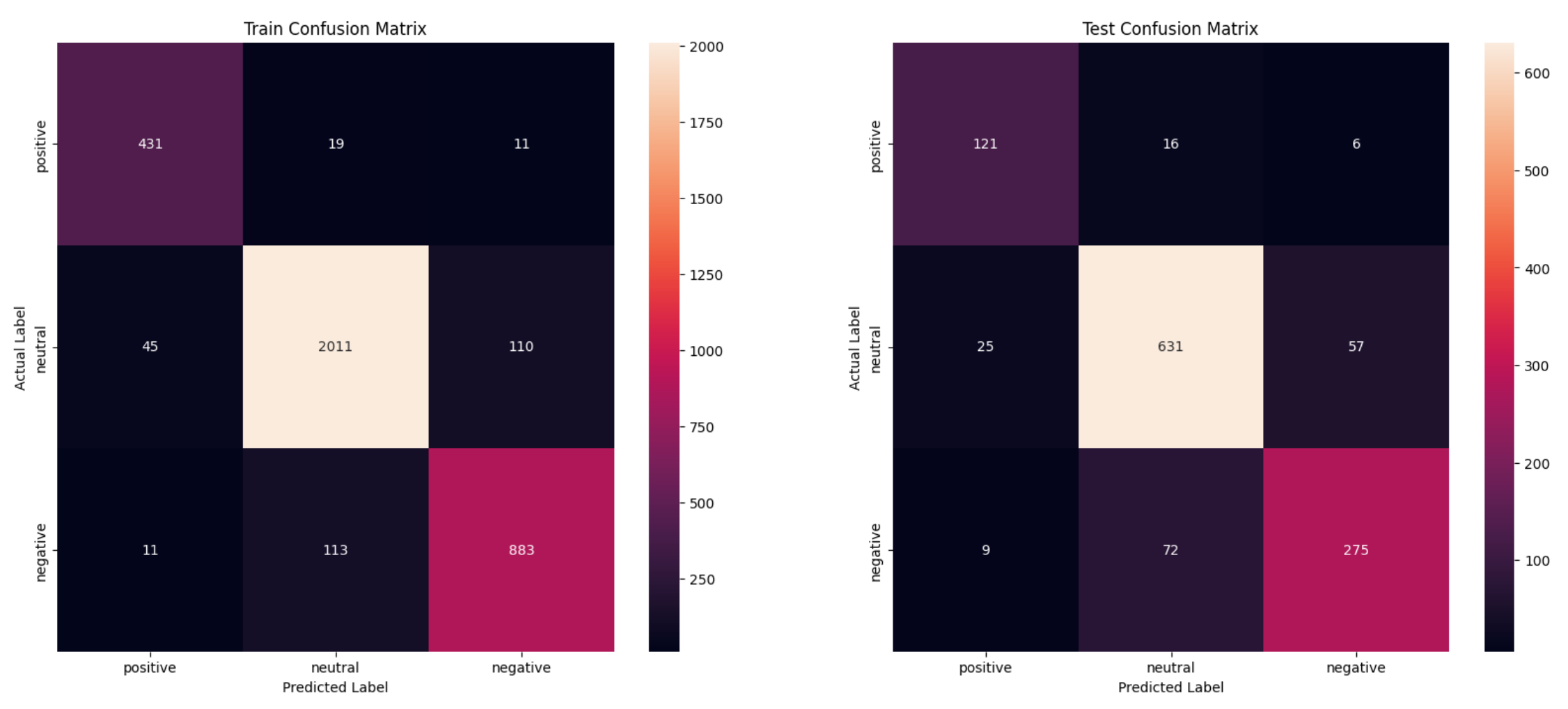}
    \caption{Confusion Matrix of Model Predictions}
    \label{fig:confusion_matrix}
\end{figure}

The confusion matrix provides a detailed breakdown of the model's predictions across all sentiment classes, as shown in Figure~\ref{fig:confusion_matrix}. This analysis helps in identifying common misclassifications and understanding the model's strengths and weaknesses.

\subsection{Training and Validation Performance}

\begin{figure}[h]
    \centering
    \includegraphics[width=0.5\textwidth]{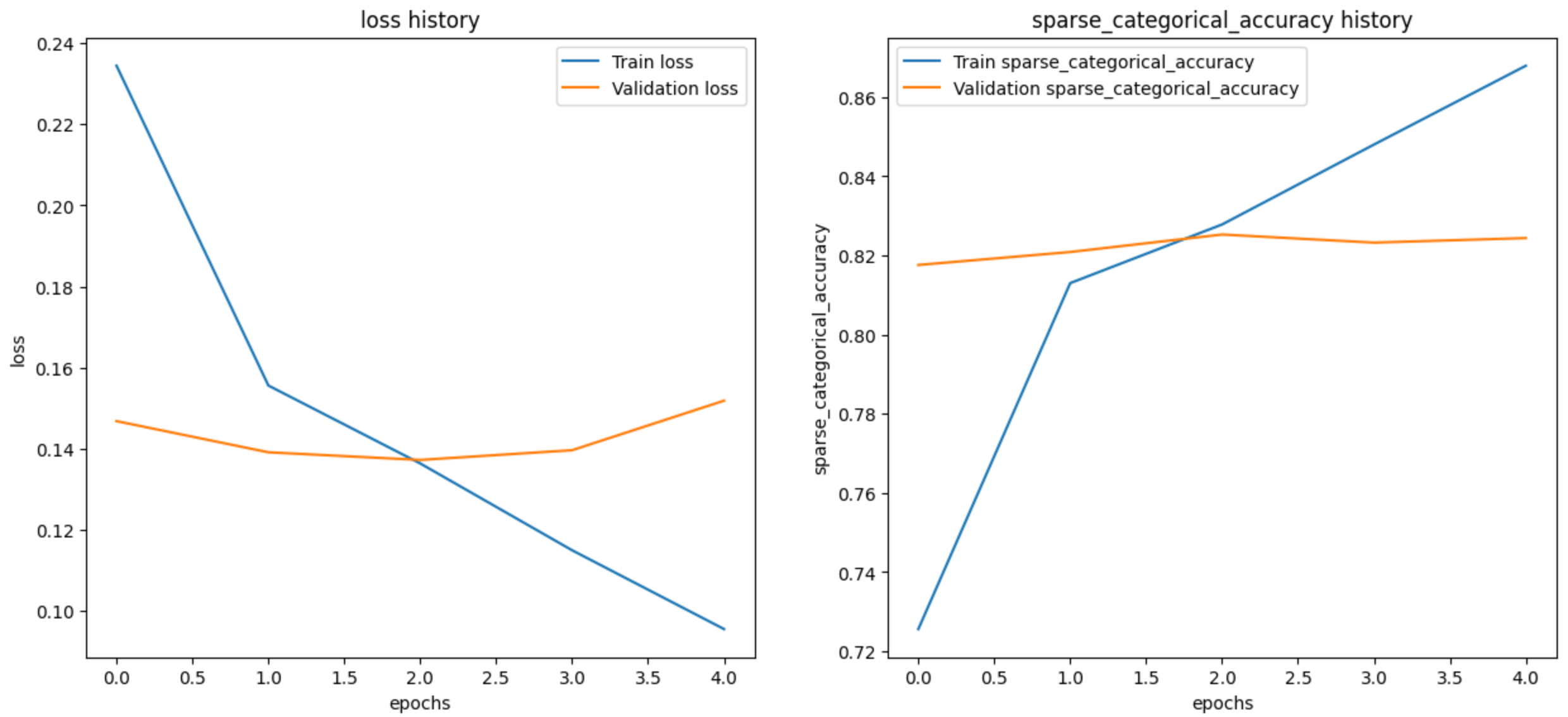}
    \caption{Train and Validation Loss and Sparse Categorical Accuracy Across Epochs}
    \label{fig:train_validation_accuracy}
\end{figure}

Figure~\ref{fig:train_validation_accuracy} illustrates the training and validation loss and sparse categorical accuracy over the epochs. The graphs show a clear convergence of the model, indicating effective training and minimal overfitting.

\subsection{Comparison with Baseline}

The performance of the fine-tuned Gemma-7B model was compared with several other models, including the baseline models and other fine-tuned models. Table~\ref{tab:comparison} highlights the precision, recall, and F1-score for each model.

\begin{table}[h]
    \centering
    \caption{Comparison of Model Performance}
    \begin{tabular}{|c|c|c|c|}
        \hline
        \textbf{Model} & \textbf{Precision} & \textbf{Recall} & \textbf{F1-score} \\
        \hline
        Bert & 0.822 & 0.824 & 0.811 \\
        \hline
        distilbert-base-uncased finetuning & 0.824 & 0.835 & 0.829 \\
        \hline
        Fine-tune Llama & 0.872 & 0.871 & 0.872 \\
        \hline
        Fine-tune Phi-3 & 0.870 & 0.861 & 0.861 \\
        \hline
        Fine-tune gemma-7b & 0.874 & 0.875 & 0.876 \\
        \hline
    \end{tabular}
    \label{tab:comparison}
\end{table}

\subsection{Error Analysis}

An error analysis was conducted to identify common patterns in misclassifications. The most frequent errors were observed in distinguishing between neutral and slightly positive or slightly negative sentiments. This indicates that the model sometimes struggles with headlines that have subtle sentiment cues.

By examining these errors, further improvements can be made to the model, such as incorporating additional context or using more advanced sentiment analysis techniques.

\section{Conclusion}

This study demonstrated the effectiveness of fine-tuning the Gemma-7B model for sentiment analysis of financial news headlines. The fine-tuning process significantly improved the model's performance, achieving an overall accuracy of 0.874, with strong precision, recall, and F1-score metrics across positive, neutral, and negative sentiments.

The model's ability to accurately classify sentiments is critical for applications in financial market analysis, risk management, and investment decision-making. Future work will focus on incorporating more diverse datasets, exploring advanced fine-tuning techniques, and integrating additional contextual information to further enhance the model's capabilities. As demonstrated in the study \cite{li2023detection}, integrating inputs from various linguistic and physiological modules significantly enhanced model performance. Thus, in our future work, we will incorporate inputs from multiple sources to further enhance our model.

In conclusion, the fine-tuned Gemma-7B model offers a powerful tool for sentiment analysis in the financial domain, with significant potential for further advancements and applications.

\bibliographystyle{IEEEtran}
\bibliography{references}

\begin{thebibliography}{10}
\providecommand{\url}[1]{#1}
\csname url@samestyle\endcsname
\providecommand{\newblock}{\relax}
\providecommand{\bibinfo}[2]{#2}
\providecommand{\BIBentrySTDinterwordspacing}{\spaceskip=0pt\relax}
\providecommand{\BIBentryALTinterwordstretchfactor}{4}
\providecommand{\BIBentryALTinterwordspacing}{\spaceskip=\fontdimen2\font plus
\BIBentryALTinterwordstretchfactor\fontdimen3\font minus \fontdimen4\font\relax}
\providecommand{\BIBforeignlanguage}[2]{{%
\expandafter\ifx\csname l@#1\endcsname\relax
\typeout{** WARNING: IEEEtran.bst: No hyphenation pattern has been}%
\typeout{** loaded for the language `#1'. Using the pattern for}%
\typeout{** the default language instead.}%
\else
\language=\csname l@#1\endcsname
\fi
#2}}
\providecommand{\BIBdecl}{\relax}
\BIBdecl

\bibitem{tetlock2007giving}
P.~C. Tetlock, ``Giving content to investor sentiment: The role of media in the stock market,'' \emph{The Journal of finance}, vol.~62, no.~3, pp. 1139--1168, 2007.

\bibitem{loughran2011liability}
T.~Loughran and B.~McDonald, ``When is a liability not a liability? textual analysis, dictionaries, and 10-ks,'' \emph{The Journal of finance}, vol.~66, no.~1, pp. 35--65, 2011.

\bibitem{xia2023parameterized}
Y.~Xia, S.~Liu, Q.~Yu, L.~Deng, Y.~Zhang, H.~Su, and K.~Zheng, ``Parameterized decision-making with multi-modal perception for autonomous driving,'' \emph{arXiv preprint arXiv:2312.11935}, 2023.

\bibitem{bollen2011twitter}
J.~Bollen, H.~Mao, and X.~Zeng, ``Twitter mood predicts the stock market,'' \emph{Journal of computational science}, vol.~2, no.~1, pp. 1--8, 2011.

\bibitem{malo2014good}
P.~Malo, A.~Sinha, P.~Korhonen, J.~Wallenius, and P.~Takala, ``Good debt or bad debt: Detecting semantic orientations in economic texts,'' \emph{Journal of the Association for Information Science and Technology}, vol.~65, no.~4, pp. 782--796, 2014.

\bibitem{pang2008opinion}
B.~Pang, L.~Lee \emph{et~al.}, ``Opinion mining and sentiment analysis,'' \emph{Foundations and Trends{\textregistered} in information retrieval}, vol.~2, no. 1--2, pp. 1--135, 2008.

\bibitem{si2013exploiting}
J.~Si, A.~Mukherjee, B.~Liu, Q.~Li, H.~Li, and X.~Deng, ``Exploiting topic based twitter sentiment for stock prediction,'' in \emph{Proceedings of the 51st Annual Meeting of the Association for Computational Linguistics (Volume 2: Short Papers)}, 2013, pp. 24--29.

\bibitem{devlin2018bert}
J.~Devlin, M.-W. Chang, K.~Lee, and K.~Toutanova, ``Bert: Pre-training of deep bidirectional transformers for language understanding,'' \emph{arXiv preprint arXiv:1810.04805}, 2018.

\bibitem{hu2004mining}
M.~Hu and B.~Liu, ``Mining and summarizing customer reviews,'' in \emph{Proceedings of the tenth ACM SIGKDD international conference on Knowledge discovery and data mining}, 2004, pp. 168--177.

\bibitem{xing2018natural}
F.~Z. Xing, E.~Cambria, and R.~E. Welsch, ``Natural language based financial forecasting: a survey,'' \emph{Artificial Intelligence Review}, vol.~50, no.~1, pp. 49--73, 2018.

\bibitem{kogan2009predicting}
S.~Kogan, D.~Levin, B.~R. Routledge, J.~S. Sagi, and N.~A. Smith, ``Predicting risk from financial reports with regression,'' in \emph{Proceedings of human language technologies: the 2009 annual conference of the North American Chapter of the Association for Computational Linguistics}, 2009, pp. 272--280.

\bibitem{barber2008all}
B.~M. Barber and T.~Odean, ``All that glitters: The effect of attention and news on the buying behavior of individual and institutional investors,'' \emph{The review of financial studies}, vol.~21, no.~2, pp. 785--818, 2008.

\bibitem{zhang2011predicting}
X.~Zhang, H.~Fuehres, and P.~A. Gloor, ``Predicting stock market indicators through twitter “i hope it is not as bad as i fear”,'' \emph{Procedia-Social and Behavioral Sciences}, vol.~26, pp. 55--62, 2011.

\bibitem{li2010information}
F.~Li, ``The information content of forward-looking statements in corporate filings—a na{\"\i}ve bayesian machine learning approach,'' \emph{Journal of accounting research}, vol.~48, no.~5, pp. 1049--1102, 2010.

\bibitem{choi2023not}
B.-G. Choi, J.~H. Choi, and S.~Malik, ``Not just for investors: The role of earnings announcements in guiding job seekers,'' \emph{Journal of Accounting and Economics}, vol.~76, no.~1, p. 101588, 2023.

\bibitem{kumar2016survey}
B.~S. Kumar and V.~Ravi, ``A survey of the applications of text mining in financial domain,'' \emph{Knowledge-Based Systems}, vol. 114, pp. 128--147, 2016.

\bibitem{mei2024efficiency}
T.~Mei, Y.~Zi, X.~Cheng, Z.~Gao, Q.~Wang, and H.~Yang, ``Efficiency optimization of large-scale language models based on deep learning in natural language processing tasks,'' \emph{arXiv preprint arXiv:2405.11704}, 2024.

\bibitem{deng2023long}
T.~Deng, H.~Xie, J.~Wang, and W.~Chen, ``Long-term visual simultaneous localization and mapping: Using a bayesian persistence filter-based global map prediction,'' \emph{IEEE Robotics \& Automation Magazine}, vol.~30, no.~1, pp. 36--49, 2023.

\bibitem{deng2024neslam}
T.~Deng, Y.~Wang, H.~Xie, H.~Wang, J.~Wang, D.~Wang, and W.~Chen, ``Neslam: Neural implicit mapping and self-supervised feature tracking with depth completion and denoising,'' \emph{arXiv preprint arXiv:2403.20034}, 2024.

\bibitem{liu2024adaptive100}
H.~Liu, Y.~Shen, W.~Zhou, Y.~Zou, C.~Zhou, and S.~He, ``Adaptive speed planning for unmanned vehicle based on deep reinforcement learning,'' \emph{arXiv preprint arXiv:2404.17379}, 2024.

\bibitem{shen2024localization}
Y.~Shen, H.~Liu, X.~Liu, W.~Zhou, C.~Zhou, and Y.~Chen, ``Localization through particle filter powered neural network estimated monocular camera poses,'' \emph{arXiv preprint arXiv:2404.17685}, 2024.

\bibitem{zhang2024deepgi}
Y.~Zhang, Y.~Gong, D.~Cui, X.~Li, and X.~Shen, ``Deepgi: An automated approach for gastrointestinal tract segmentation in mri scans,'' \emph{arXiv preprint arXiv:2401.15354}, 2024.

\bibitem{liu2024enhanced}
R.~Liu, X.~Xu, Y.~Shen, A.~Zhu, C.~Yu, T.~Chen, and Y.~Zhang, ``Enhanced detection classification via clustering svm for various robot collaboration task,'' \emph{arXiv preprint arXiv:2405.03026}, 2024.

\bibitem{zhang2024development}
Y.~Zhang, M.~Zhu, K.~Gui, J.~Yu, Y.~Hao, and H.~Sun, ``Development and application of a monte carlo tree search algorithm for simulating da vinci code game strategies,'' \emph{arXiv preprint arXiv:2403.10720}, 2024.

\bibitem{zhu2024ensemble}
M.~Zhu, Y.~Zhang, Y.~Gong, K.~Xing, X.~Yan, and J.~Song, ``Ensemble methodology: Innovations in credit default prediction using lightgbm, xgboost, and localensemble,'' \emph{arXiv preprint arXiv:2402.17979}, 2024.

\bibitem{yuan2024research}
J.~Yuan, L.~Wu, Y.~Gong, Z.~Yu, Z.~Liu, and S.~He, ``Research on intelligent aided diagnosis system of medical image based on computer deep learning,'' \emph{arXiv preprint arXiv:2404.18419}, 2024.

\bibitem{liu2024spam}
T.~Liu, S.~Li, Y.~Dong, Y.~Mo, and S.~He, ``Spam detection and classification based on distilbert deep learning algorithm,'' \emph{Applied Science and Engineering Journal for Advanced Research}, vol.~3, no.~3, pp. 6--10, 2024.

\bibitem{li2023detection}
P.~Li, M.~Abouelenien, R.~Mihalcea, Z.~Ding, Q.~Yang, and Y.~Zhou, ``Deception detection from linguistic and physiological data streams using bimodal convolutional neural networks,'' \emph{arXiv preprint arXiv:2311.10944}, 2023.

\end{thebibliography}
\end{CJK*}	
\end{document}